%% file: main.tex
\documentclass{article}
\usepackage{iclr2027_conference,times}

\PassOptionsToPackage{table}{xcolor}
\usepackage{amsmath,amssymb,amsthm}
\usepackage{mathtools}
\usepackage{graphicx}
\usepackage{booktabs}
\usepackage{array}
\newcolumntype{L}[1]{>{\raggedright\arraybackslash}p{#1}}
\usepackage{multirow}
\usepackage{xcolor}
\usepackage{colortbl}
\usepackage{hyperref}
\usepackage{xspace}
\definecolor{refgreen}{HTML}{0B7A3B}
\hypersetup{colorlinks=true, citecolor=blue, linkcolor=refgreen, urlcolor=blue}
\newcommand{\secref}[1]{\hyperref[#1]{Sec.~\ref*{#1}}}
\newcommand{\appref}[1]{\hyperref[#1]{Appendix~\ref*{#1}}}
\newcommand{\tabref}[1]{\hyperref[#1]{Table~\ref*{#1}}}
\newcommand{\figref}[1]{\hyperref[#1]{Fig.~\ref*{#1}}}
\newcommand{\eqnref}[1]{\hyperref[#1]{Eq.~(\ref*{#1})}}
\newcommand{\lemref}[1]{\hyperref[#1]{Lemma~\ref*{#1}}}
\usepackage{url}

\theoremstyle{remark}

\theoremstyle{plain}

\newcommand{\model}{\textsc{Loci}}
\definecolor{ourrow}{HTML}{FDF1EA}   %

\title{LOCI: Spatial Linear Memory\\for Streaming World Models}

\iclrfinalcopy

\author{Ji Xia$^{1}$ \quad Tingting Liao$^{1}$ \quad Xuezhi Liang$^{1}$ \quad Hao Li$^{2,3}$ \quad Guangyi Liu$^{1\dagger}$ \\
$^{1}$Institute of Foundation Models, Mohamed bin Zayed University of Artificial Intelligence \\
$^{2}$Mohamed bin Zayed University of Artificial Intelligence \quad $^{3}$Pinscreen}

\makeatletter
\newenvironment{teaser}{\par\vspace{-20pt}\begin{center}\small\let\normalsize\small\def\@captype{figure}}{\end{center}\vspace{-6pt}}
\makeatother

\begin{document}

\maketitle
\lhead{}\renewcommand{\headrulewidth}{0pt}
{\renewcommand{\thefootnote}{}\footnotetext{$^{\dagger}$Corresponding author.}}

\begin{teaser}
\includegraphics[width=\linewidth]{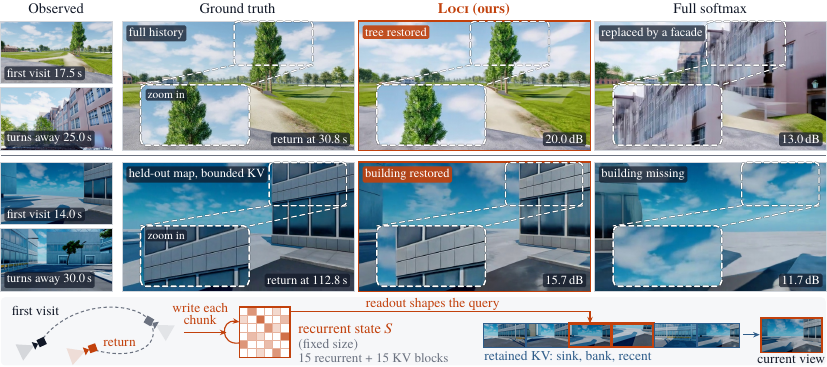}
\caption{\textbf{Returning to a previously seen place.} Each model continues the recorded video (top: full history; bottom: held-out Tokyo map, same bounded KV budget). Full softmax, trained with the same recipe, replaces the tree with a facade and loses the building; \model{} restores both (PSNR against ground truth; insets magnify the boxed region). Band: a fixed-size recurrent state shapes the queries into the retained KV (thumbnails: frames the Tokyo run retains).}
\label{fig:teaser}
\end{teaser}

\input{sections/00_abstract}
\input{sections/01_intro}

\input{sections/02_related}

\input{sections/03_preliminaries}
\input{sections/04_method}
\input{sections/06_experiments}
\input{sections/07_discussion}

\newpage
\section*{Acknowledgments}
We thank Xin Li and Fan Yang for their valuable help with the manuscript and figures, and for insightful discussions.

\bibliography{refs}
\bibliographystyle{iclr2027_conference}

\appendix
\input{sections/08_appendix}
\input{sections/09_appendix_protocol}

\end{document}

%% file: sections/00_abstract.tex
\begin{abstract} 
When a camera revisits a previously observed region, a video world model should reproduce what was there before. This requires both remembering past observations and retrieving the right one for the current viewpoint. Key--value caches preserve visual detail but grow with video length; recurrent memory is compact but compresses history into a fixed-size state, so individual past observations are no longer directly accessible. We introduce LOCI, a hybrid spatial-memory architecture that keeps both representations. In half of the transformer blocks, main attention keeps a key--value cache of past observations; in the other half, it is restricted to the current chunk and complemented by a recurrent linear-attention memory whose reads and writes are conditioned on projective camera geometry, so viewpoint enters both memory addressing and stored content. Recurrent readouts flow into subsequent cache-backed blocks and supply their queries with accumulated scene context. On the public MIND memory benchmark and on held-out recorded trajectories, LOCI reproduces revisited content more faithfully than representative world models and a same-recipe full-softmax model; with full history, it lowers peak memory at equal length by about 30\% relative to full softmax. With a bounded bank of retained observations, it streams long videos at constant memory and remains more faithful than full softmax under the same budget.
Project page: \url{https://xiaji2021.github.io/LOCI/}.

\end{abstract}

%% file: sections/01_intro.tex
\section{Introduction}

Camera-controllable video world models now generate long, interactive explorations from actions or camera trajectories \citep{hyworld15,matrixgame3,lingbotworld,zhu2026sanawm,chen2026reworld}. Beyond plausible local continuations, such a model must preserve scene structure and appearance when the camera revisits observed regions, even after a prolonged absence. This \emph{spatial persistence} requires both accurate camera control and historical evidence: the requested viewpoint determines where to look, while previous observations constrain what should appear. Streaming generation is challenging because spatial relevance does not follow temporal recency; an observation outside the recent context may be essential to reconstructing the current view.

Historical attention preserves individually accessible key--value (KV) features, but full-history storage and access costs grow with trajectory length. Selecting historical observations or pose-indexed landmarks reduces the active attention budget \citep{yu2025contextasmemory,xiao2025worldmem,li2025vmem,xu2026wonder,chen2026reworld} but can exclude evidence needed later. Local--global architectures combine fixed-size recurrent states with detailed local attention \citep{li2026hybridforcing,zhu2026sanawm}, but consolidating observations into a state can lose scene-specific detail; as \citet{xue2026ring} observe, long-horizon consistency needs both persistent memory and a long context. Retaining historical KV preserves that detail without guaranteeing its use: the current query must still address relevant evidence, and generation must incorporate it. We keep both traces of the past and let them interact. A camera-conditioned recurrent state summarizes the entire history at fixed size, while retained KV preserves observation-level evidence. Because the recurrent readout is added to the token features from which later layers form their queries, \emph{accumulated scene context can direct attention in the retained KV toward observations relevant to the current view}.

\begin{figure}[t]
\centering
\includegraphics[width=0.98\linewidth]{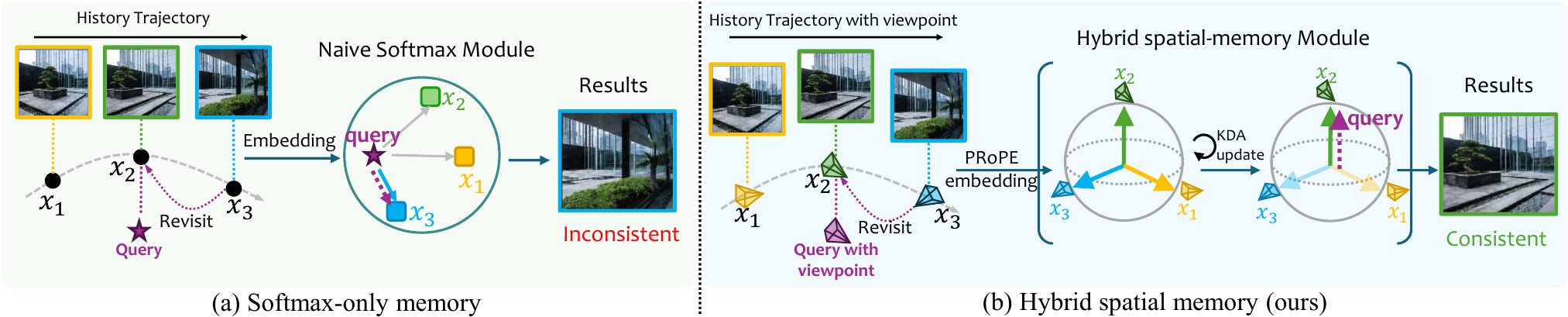}
\caption{Key idea of \model{}. Projectively conditioned  recurrent memory integrates historical context, while retained KV preserves observation-level detail. Recurrent readouts update the features used by subsequent historical attention, connecting the two representations through the feature stream.}
\label{fig:keyidea}
\end{figure}

We introduce \model{}, a hybrid spatial-memory architecture built around this interaction (\figref{fig:keyidea}). We adopt the layer layout of ARL$^2$~\citep{li2026arl2}, a text-to-video hybrid, interleaving blocks that combine intra-chunk attention with recurrent memory and blocks that retain historical softmax attention. In each hybrid block, current tokens read the state established by preceding chunks. A learned gate scales the recurrent readout before it is added to the local-attention output. The resulting features propagate through the network, and subsequent historical-attention blocks construct their queries from them. 
Recurrent context informs these queries under both full and bounded history. With a bounded bank, the recurrent state additionally carries information from observations whose KV entries have been discarded.

Built on Kimi Delta Attention \citep{kimilinear2025}, the recurrent path has two adaptations for video scene memory. First, PRoPE \citep{prope2025} conditions queries, keys, and values on camera geometry, so what the recurrent memory writes and reads depends on viewpoint. Second, token-level delta corrections revise individual associations, while channel-wise retention is applied once per chunk. This decouples explicit forgetting frequency from spatial token count without coarsening associative updates.

With a 5B backbone, we evaluate fidelity to the recorded video at revisits against representative world models and a full-softmax baseline trained with the same backbone, data, recipe and number of updates. Relative to this baseline, \model{} improves reference PSNR at revisits by 0.62 dB on held-out Unreal Engine trajectories, and by 0.99 dB on set A when both models access the same bounded set of retained observations. On the public MIND benchmark, it is significantly better over full prediction segments under the same bounded budget (PSNR +0.89 dB). Our \textbf{contributions} are threefold:
\begin{itemize}
\item We develop a hybrid spatial memory that couples projectively conditioned recurrent integration with direct historical KV access, allowing recurrent context to inform historical queries while preserving observation-level detail. Channel-wise retention is applied once per chunk, so explicit forgetting follows elapsed video time rather than token order; in an extended comparison on MIND, this lowers local error at revisits more than 20\,s apart relative to per-token retention (\appref{app:retention_ablation}).

\item We pair the recurrent state, carried over the full history, with a fixed-capacity bank of retained observations for bounded streaming. Under an identical bounded KV budget, the recurrent path improves fidelity over a same-recipe full-softmax model on the MIND memory test (all 50 segments; full-segment PSNR $+0.89$\,dB, 95\% interval $[+0.64, +1.15]$, with lower LPIPS and MSE), while generation runs for 300 seconds at constant memory.

\item On the public MIND memory benchmark \citep{mind2026}, \model{} with full-history access, evaluated zero-shot, attains lower MSE and higher PSNR and SSIM than the values reported for GIM-World \citep{gimworld2026}, which is trained on MIND. At short-horizon revisits, the hybrid is also more consistent locally than full softmax.
\end{itemize}

%% file: sections/02_related.tex
\section{Related Work}

\paragraph{Memory for scene revisits.}
Explicit-memory world models select past observations for the current view by field-of-view overlap \citep{yu2025contextasmemory,xiao2025worldmem}, a surfel index \citep{li2025vmem}, or query--key similarity over cached KV \citep{xu2026wonder}, or reproject latent patches using depth \citep{matrixgame35}. ReWorld \citep{chen2026reworld} retrieves chunks from a pose-indexed landmark bank under a fixed KV budget, and the concurrent WorldCrafter \citep{yu2026worldcrafter} compresses selected views into a fixed set of target-view tokens through a pose-guided readout. Training-free methods retrieve by pose or camera similarity \citep{ma2026closing,yi2026worldkv}, curate or recall cached KV by content \citep{xu2026recap,wu2026echo}, or remap positions so that distant memory stays within the trained range, training-free \citep{wu2026addressable} or with ring-structured training \citep{xue2026ring}. LayerRecall \citep{ding2026layerrecall} trains a state-conditioned router that retrieves historical KV and injects it into a fixed set of memory-sensitive layers. Others learn implicit memory \citep{car2026,gimworld2026} or target long-horizon streaming \citep{hyworld15,matrixgame3}. \model{} also keeps observation-level KV, bounded by a bank of diverse views rather than per-target-view retrieval, and adds a recurrent state over the full history whose readout shapes downstream queries without a separate router or objective.

\paragraph{Hybrid recurrent video models.}
Several video models pair a linear-attention or state-space recurrence \citep{yang2025gateddeltanet,kimilinear2025} with local softmax attention: Hybrid Forcing \citep{li2026hybridforcing} accumulates evicted KV additively, SANA-WM \citep{zhu2026sanawm} interleaves camera-conditioned Gated DeltaNet, in which all tokens of a latent frame form one recurrent step, with sink-and-window softmax blocks, and Video SSM \citep{po2025ssm} uses a block-wise state-space scan. Remote content at a revisit is then available only through the compressed state; Astronex-World \citep{zhou2026astronex}, with the same backbone and PRoPE camera encoding, keeps only an attention sink and a fixed local window without a recurrent path. ARL$^2$ \citep{li2026arl2}, a text-to-video conversion recipe without camera control or revisit evaluation, supplies our layer layout and read-then-commit schedule; instead of converting a trained model, we train the hybrid with chunk-wise diffusion forcing. In \model{}, the recurrent memory writes token by token and applies retention once per chunk; the remaining blocks attend directly to retained observations through queries formed from features that include the camera-conditioned recurrent readout; and revisits are evaluated against recorded ground truth (\appref{app:hybrid_related}).

\paragraph{Camera conditioning.} 
Building on ray-based camera features~\citep{he2024cameractrl}, every block keeps a UCPE camera-attention branch~\citep{zhang2025ucpe}. PRoPE~\citep{prope2025} was formulated for softmax attention; we apply it inside delta-rule memory, so the state stores associations between camera-transformed keys and values. ViewRope~\citep{xiang2026viewrope} and MeRoPE~\citep{qiao2026merope} are rotary alternatives.

%% file: sections/03_preliminaries.tex
\section{Preliminaries}
\label{sec:prelim}

\paragraph{Kimi Delta Attention.}
Kimi Delta Attention (KDA) \citep{kimilinear2025} keeps a fixed-size state $S\in\mathbb{R}^{d_k\times d_v}$ through channel-wise retention and token-level delta corrections. For a normalized key $k_t$, value $v_t$, diagonal retention $D_t$, and write strength $\beta_t=2\sigma(b_t)\in(0,2)$ in our parameterization,
\begin{equation}
\widetilde S_t=D_tS_{t-1},\qquad
S_t=\widetilde S_t+\beta_t k_t
\bigl(v_t-\widetilde S_t^{\top}k_t\bigr)^{\top}.
\label{eq:kda}
\end{equation}
Retention controls which key channels persist, while the delta residual corrects the value predicted at $k_t$ \citep{yang2024deltanet,yang2025gateddeltanet}. A query $q$ reads $S^\top q$ from the available state. Our chunk-wise read-before-write schedule is specified in \secref{sec:granularity}.

\paragraph{Rotary and camera-relative positional encoding.}
RoPE \citep{su2021roformer} rotates queries and keys with $\Phi(p)$, where $\Phi(p_i)^\top\Phi(p_j)=\Phi(p_j-p_i)$ makes their interaction depend on relative position. PRoPE \citep{prope2025} extends this to camera geometry. Each token has a transform $G_i$ combining its camera projection matrix $P_i$ with patch-level RoPE:
\begin{equation}
o_i = G_i\sum_j \operatorname{softmax}_j
\bigl(q_i^\top G_iG_j^{-1}k_j\bigr)\,G_j^{-1}v_j.
\label{eq:prope}
\end{equation}
Camera dependence enters through relative projective transforms $P_iP_j^{-1}$, so this attention is invariant to a common change of world coordinates. We apply projective conditioning to recurrent memory.

%% file: sections/04_method.tex
\section{LOCI}
\label{sec:method}

\paragraph{Architecture.}
We adapt the 30-block Wan video transformer \citep{wan2025} to chunk-causal generation.
A clean conditioning latent $z_0$ is followed by chunks $z_c$ of five latent frames,
with camera pose and intrinsics supplied for each frame:
\begin{equation}
 p_\theta(z_{1:C}\mid z_0,\pi_{0:C})
 =\prod_{c=1}^{C}p_\theta(z_c\mid z_{<c},\pi_{0:c}).
 \label{eq:factorization}
\end{equation}
Fifteen hybrid blocks combine intra-chunk softmax attention with recurrent Kimi Delta
Attention (KDA) \citep{kimilinear2025}; the other fifteen retain softmax attention
with explicit historical keys and values. All blocks keep an independent UCPE
ray-conditioned camera-attention branch \citep{zhang2025ucpe} (\appref{app:impl}). In every block the camera branch attends to the same retained frames as the historical-attention blocks.
At an equal retained-frame budget, \model{} thus stores main-attention KV in 15 layers rather than 30 and replaces the other 15 histories with fixed-size states. Camera KV stays in all 30 blocks.

\begin{figure}[t]
 \centering
 \vspace{-20pt}
 \includegraphics[width=\linewidth]{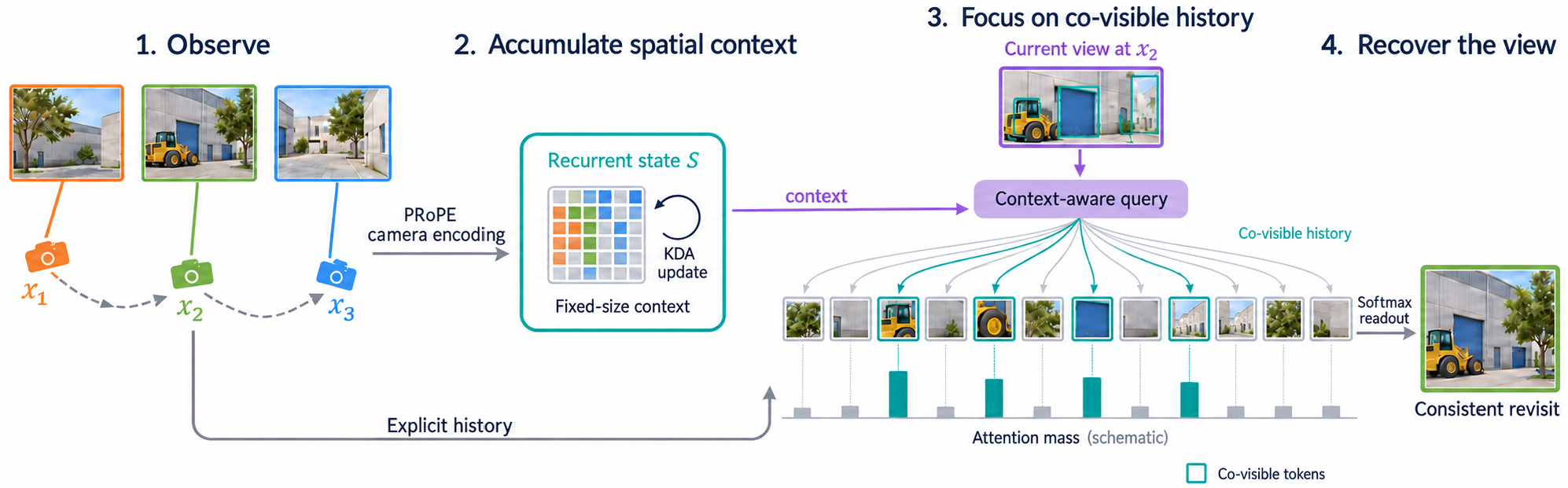}
   \vspace{-10pt}
\caption{\textbf{Conceptual illustration of \model.} A camera-conditioned recurrent state accumulates context from past observations; at a revisit, its readout shapes the queries with which later historical-attention blocks attend to co-visible retained observations. The layer layout is given in \appref{app:layout}.}
 \label{fig:loci_overview}
  \vspace{-10pt}
\end{figure}

Information follows the current chunk through network depth. At a hybrid block,
local attention describes the chunk while a PRoPE-conditioned read supplies recurrent
context from preceding chunks. Their gated sum updates the token features. A later
historical-attention block forms its queries from these features and reads explicit
keys and values from the permitted history. Compressed history therefore shapes the
representation used to access retained observations through the ordinary inter-layer
path, without a separate state-to-query adapter (\figref{fig:loci_overview}).

\subsection{Projective Camera Encoding for Recurrent Memory}
\label{sec:camrope}

A previously observed surface can become relevant after a long temporal gap, while a
recent observation may face elsewhere; camera geometry thus supplies a cue
distinct from temporal proximity. We use PRoPE \citep{prope2025} to condition the recurrent
queries, keys and values on each token's camera ray. With world-to-ray transform $E_i$
(first-frame reference, translation divided by $s=6$) and normalized intrinsics
$\overline K_i$, the projection is $P_i=\operatorname{lift}(\overline K_i)E_i$.
The query map $A_i$ applies $P_i^\top$ and the key and value maps $B_i,C_i$ apply
$P_i^{-1}$, together with patch rotations, as in \eqnref{eq:projective_maps}. The recurrent
features are
\begin{equation}
 \widehat q_i=\mathcal N(A_iq_i),\qquad
 \widehat k_i=\mathcal N(B_ik_i),\qquad
 \widetilde v_i=C_iv_i,
 \label{eq:geokeys}
\end{equation}
where $\mathcal N(x)=x/\sqrt{\|x\|_2^2+\epsilon}$ acts over the full head; the readout is
mapped back by $C_i^{-1}$. Each projective tile contributes the relative product
$q_i^\top P_iP_j^{-1}k_j$ to a query--key pairing. We disable temporal rotary encoding
only in the KDA branch. All softmax branches keep the backbone's native encoding
(\appref{app:method_details}).

\subsection{Chunk-Synchronous Memory Dynamics}
\label{sec:granularity}

\paragraph{Read from the preceding state.}
For each hybrid block and head, let $S_{c-1}\in\mathbb R^{d\times d}$ ($d=128$) be the
key-by-value state available before chunk $c$. Every token $i$ in the chunk reads this
same state, while local softmax handles interactions within the chunk:
\begin{equation}
 r_i=C_i^{-1}\!\left(d^{-1/2}S_{c-1}^{\top}\widehat q_i\right),\qquad
 o_i=W_o\!\left[o_i^{\mathrm{local}}+
       \eta\,\sigma(g_i)\,r_i\right],
 \label{eq:read}
\end{equation}
where $o_i^{\mathrm{local}}$ is intra-chunk softmax attention, $g_i$ a learned
per-token, per-head gate, and $\eta=1$ a fixed recurrent branch scale.

\paragraph{Write at token resolution.}
Starting from $S_{c-1}$, the tokens of the chunk update the state in sequence with the
delta rule, so new observations revise the value already associated with a key:
\begin{equation}
\begin{aligned}
 \overline S^{(i)}&=D_iS^{(i-1)},\qquad \beta_i=2\sigma(b_i)\\
 S^{(i)}&=\overline S^{(i)}+
 \beta_i\widehat k_i
 \left(\widetilde v_i-\overline S^{(i)\top}\widehat k_i\right)^\top .
\end{aligned}
\label{eq:geokda}
\end{equation}
The scan's final state becomes $S_c$: queries read the preceding chunk's state,
while writes retain token resolution \citep{li2026arl2}.

\paragraph{Retention once per chunk.}
Per-token retention, as in KDA for language, makes forgetting a function of raster
position rather than elapsed time: tokens of the same frame, observed simultaneously, are
attenuated unequally, and mostly the final tokens of a chunk survive in the state. We
therefore apply learned diagonal retention $D_i$ only at the first token of each chunk,
computed from the chunk's mean representation, and identity retention elsewhere;
see \eqnref{eq:chunkdecay}. Explicit forgetting then advances once per chunk, while delta
corrections remain token-wise (derivation in \appref{app:method_details}; comparison with per-token retention in \appref{app:retention_ablation}).

\paragraph{Training and sampling.}
Training uses chunk-wise diffusion forcing \citep{chen2024diffusionforcing}: only the
conditioning latent is clean, and a single full-window forward scans the noised chunks
from zero state. During sampling, following ARL$^2$ \citep{li2026arl2}, every denoising
iteration reads the committed state without modifying it. After the last denoising step of the current chunk, a separate forward on that sample commits the state for the next chunk
(timestep zero with full history; $t=100$ in the bounded sparse setting).
\appref{app:state_schedule} summarizes both schedules.

\subsection{Dense and Bounded Sparse Historical Access}
\label{sec:retrieval}

A fixed-size state compresses context; explicit KV preserves selected token-level detail
for direct attention. In the dense mode, historical softmax blocks attend over the
evaluated prefix. To bound explicit storage during streaming, the sparse mode retains a
recent window and a diverse bank of older views, shared by the main and camera branches:
\begin{equation}
 \mathcal F_c=\{0\}\cup\mathcal B_c\cup\mathcal R_c\cup\mathcal U_c,
 \qquad |\mathcal B_c|\le20,\quad |\mathcal R_c|\le8,\quad |\mathcal U_c|=5.
 \label{eq:recall}
\end{equation}
The conditioning frame is a sink, $\mathcal R_c$ holds the most recent completed frames,
and $\mathcal U_c$ is the current chunk, so softmax attention sees at most 34 latent
frames. The panorama bank $\mathcal B_c$ is updated from its previous contents and the
frames leaving the recent window by a field-of-view coverage criterion that favors
earlier observations adding complementary coverage (\appref{app:impl}). Discarded
observations are not archived. Bank selection leaves the recurrent state intact, so
the fixed-size state and bounded bank let generation continue without growing either
history store. We analyze storage and per-chunk cost in
\appref{app:method_details}.

%% file: sections/06_experiments.tex
\section{Experiments}
\label{sec:exp}

\subsection{Experimental setup}
\label{sec:setup}

\paragraph{Models and baselines.}
All our models start from Wan2.2-TI2V-5B \citep{wan2025} and are trained for 5{,}000 updates with chunk-wise diffusion
forcing. \emph{Full softmax} keeps softmax attention in all 30 blocks under the identical recipe, data and number of
updates. The recurrent branch adds 90.7M parameters (1.7\% of 5.38B). Inference uses 50 denoising steps and CFG 1. We run external camera-controlled world models with official
weights and default sampling: CaR \citep{car2026}, HY-WorldPlay \citep{hyworld15}, Matrix-Game~3.0 \citep{matrixgame3},
LingBot-World \citep{lingbotworld}, AlayaWorld \citep{alayaworld2026}, Alaya-EVOKE \citep{evoke2026}, SANA-WM \citep{zhu2026sanawm} and
SolarWM \citep{solarwm2026}. ARL$^2$ \citep{li2026arl2} is reproduced with its conversion recipe. From update
2{,}000, our main models are trained on a UE-weighted data mixture; a second pair continues on the uniform mixture.
Training uses Unreal Engine and CARLA scenes that we rendered (about 98\,h) and real walking videos from Sekai
\citep{li2025sekai}. No MIND data is used. Details are in \appref{app:bench}.

\paragraph{Evaluation.}
The main benchmark is MIND \citep{mind2026}: a memory segment is supplied as context and the continuation is compared
with the recording. We follow its official protocol and evaluate the entire prediction segment; results over the first
20\,s are in \appref{app:mind20}. We list the numbers reported by GIM-World \citep{gimworld2026}, which is trained on MIND,
as a reference, and evaluate \model{} zero-shot. We also report WBench's gated camera-return consistency
\citep{wbench2026} and held-out recorded Unreal Engine trajectories with ground truth (\appref{app:recorded}). A
\emph{revisit instant} is a pose within 0.15\,m and $32^\circ$ of an earlier one, reached after at least 8\,s and after
looking away by $60^\circ$. Metrics are MSE, PSNR, SSIM \citep{wang2004ssim} and LPIPS \citep{zhang2018lpips}. Intervals are 95\% bootstrap intervals over
clips (10{,}000 resamples).

\subsection{Comparison with SOTA}
\label{sec:main}

\label{sec:thirdparty}
\begin{figure}[t]
\centering
\includegraphics[width=\linewidth]{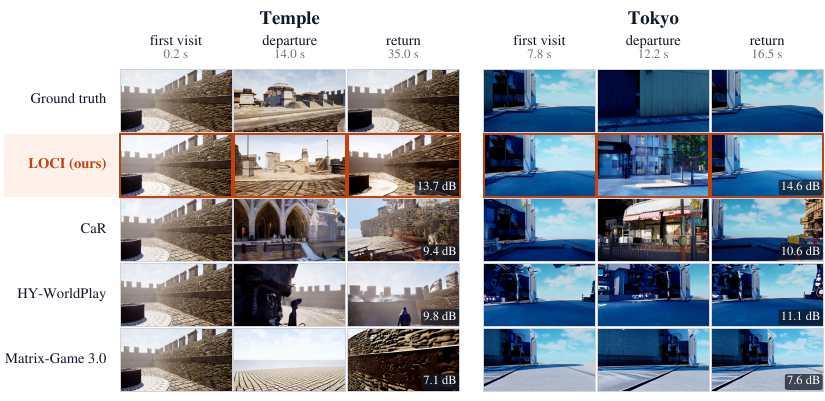}
\caption{\textbf{Qualitative revisit comparison} on held-out recorded trajectories (first frame and camera trajectory as
input for all methods). Each row is one model; columns show the first visit, a view after the camera has turned away,
and the return to the first viewpoint, with the PSNR of the return frame against ground truth. \model{} restores the
previously observed scene on return, whereas the other models replace or distort it. For each clip we show the return
at which \model{} leads by the largest margin; clip-level averages are in \tabref{tab:sota}. Temple Plaza is a training map traversed along a new trajectory; the Tokyo map is not used in training.}
\label{fig:revisit_results}
\end{figure}

External models differ in size, training data, sampling steps and how they receive the memory segment
(\appref{app:mindadapters}), so this comparison is a reference under a unified protocol; the controlled comparison in
\secref{sec:ablation} isolates the architecture.
On MIND (\tabref{tab:thirdparty}), \model{} obtains lower MSE and higher PSNR and SSIM than reported for
GIM-World, which is trained on the benchmark's data, and for Context-as-Memory, FramePack and SSM,
with LPIPS below all of them except GIM-World (0.643 vs.\ 0.630). Among the streaming world models we run with the memory segment as context, several of them larger (8--28B),
\model{} is best on all four metrics among those scored on all segments, 1.82\,dB PSNR above the strongest
(HY-WorldPlay). With bounded sparse history at constant memory, it still has the lowest MSE and the highest PSNR and
SSIM of this group. On the 36 segments that CaR completes, \model{} has lower MSE and higher PSNR and SSIM,
whereas CaR has lower LPIPS (0.622 vs.\ 0.630).
\tabref{tab:thirdparty2} (\appref{app:extra}) reports WBench-Navi gated camera-return consistency \citep{wbench2026}.
On the held-out recorded trajectories of sets A and B (\appref{app:recorded}), the ranking holds: at revisit
instants \model{} has significantly lower LPIPS than six of the seven external models we run and significantly higher
PSNR than five of them (\tabref{tab:paired}; examples in \figref{fig:revisit_results}).

\begin{table}[t]
  \vspace{-10pt}
\caption{\textbf{MIND memory test} (first-person). \emph{Reference}: numbers reported in the GIM-World paper, also
evaluated on the entire prediction segment (not re-run by us). \emph{Run by us}: all models evaluated identically on the entire prediction segment
with the official evaluation code; external models use official weights and
default sampling, with the memory segment provided as context. $n$: number of scored segments (Matrix-Game~3.0: one
segment failed in the official scorer; LingBot-World: 28B in total, 14B active). $^\ddagger$CaR: the 36 shortest segments; generation of the 14 longest did not finish
(on these 36 segments \model{} obtains 0.0460 / 14.39 / 0.478 / 0.630). Bold: best among models run by us on all segments.}
\label{tab:thirdparty}
\centering\scriptsize
\setlength{\tabcolsep}{2.75pt}
\renewcommand{\arraystretch}{1.1}
\begin{tabular}{l c c c cccc}
\toprule
Method & Params & Trained on MIND & $n$ & MSE$\downarrow$ & PSNR$\uparrow$ & SSIM$\uparrow$ & LPIPS$\downarrow$ \\
\midrule
\multicolumn{8}{@{}l}{\emph{Reference (reported in the GIM-World paper)}} \\
SSM \citep{po2025ssm} & -- & \checkmark & -- & 0.0796 & 11.96 & 0.395 & 0.744 \\
FramePack \citep{framepack2025} & -- & \checkmark & -- & 0.0764 & 12.04 & 0.376 & 0.706 \\
Context-as-Memory \citep{yu2025contextasmemory} & -- & \checkmark & -- & 0.0706 & 12.50 & 0.392 & 0.695 \\
GIM-World \citep{gimworld2026} & 1.3B & \checkmark & -- & 0.0614 & 13.40 & 0.414 & 0.630 \\
\midrule
\multicolumn{8}{@{}l}{\emph{Run by us (entire prediction segment)}} \\
HY-WorldPlay \citep{hyworld15}       & 8B  & $\times$ & 50 & 0.0710 & 12.54 & 0.398 & 0.715 \\
Matrix-Game 3.0 \citep{matrixgame3}  & 5B  & $\times$ & 49 & 0.0866 & 11.69 & 0.373 & 0.666 \\
AlayaWorld \citep{alayaworld2026}   & 15B & $\times$ & 50 & 0.0908 & 11.33 & 0.358 & 0.683 \\
Alaya-EVOKE \citep{evoke2026}  & 14B & $\times$ & 50 & 0.0884 & 11.14 & 0.347 & 0.710 \\
LingBot-World \citep{lingbotworld}   & 28B & $\times$ & 50 & 0.0955 & 11.33 & 0.331 & 0.703 \\
CaR$^\ddagger$ \citep{car2026}        & 5B  & $\times$ & 36 & 0.0538 & 13.85 & 0.458 & 0.622 \\
\rowcolor{ourrow} \textbf{\model{} (ours)}       & 5B & $\times$ & 50 & \textbf{0.0455} & \textbf{14.36} & \textbf{0.464} & \textbf{0.643} \\
\rowcolor{ourrow} \model{} (bounded sparse)      & 5B & $\times$ & 50 & 0.0622 & 13.02 & 0.409 & 0.679 \\
\bottomrule
\end{tabular}
\end{table}

\subsection{Ablation study}
\label{sec:ablation}

\begin{table}[t]
\caption{\textbf{Hybrid vs.\ full softmax with the same backbone, data, recipe and 5{,}000 updates} on the MIND memory
test (entire prediction segment; $\Delta$ is the paired mean over segments, \model{} $-$ full softmax, with 95\% bootstrap intervals).
Bounded: both models access the same fixed set of retained frames (first frame, bank of 20, 8 most recent).
Full history: the 32 segments completed by both models (full softmax, run on the 39 shorter segments, exceeds GPU memory
on 7 of them).}
\label{tab:ablation}
\centering\scriptsize
\setlength{\tabcolsep}{3.25pt}
\renewcommand{\arraystretch}{1.1}
\begin{tabular}{ll c cc c cc c}
\toprule
& & & \multicolumn{3}{c}{\textbf{PSNR}$\uparrow$} & \multicolumn{3}{c}{\textbf{LPIPS}$\downarrow$} \\
\cmidrule(lr){4-6}\cmidrule(l){7-9}
Data mixture & History & $n$ & Softmax & \model{} & $\Delta$ [95\% CI] & Softmax & \model{} & $\Delta$ [95\% CI] \\
\midrule
Uniform     & bounded & 50 & 12.17 & \textbf{12.85} & $+0.67$ [$+0.42$, $+0.93$] & 0.700 & \textbf{0.680} & $-0.020$ [$-0.028$, $-0.011$] \\
UE-weighted & bounded & 50 & 12.13 & \textbf{13.02} & $+0.89$ [$+0.64$, $+1.15$] & 0.704 & \textbf{0.679} & $-0.024$ [$-0.034$, $-0.015$] \\
UE-weighted & full    & 32 & 13.90 & \textbf{14.25} & $+0.35$ [$+0.08$, $+0.62$] & 0.669 & \textbf{0.641} & $-0.029$ [$-0.040$, $-0.018$] \\
\bottomrule
\end{tabular}
\end{table}

\paragraph{Full softmax vs.\ hybrid.}
With identical data, recipe and number of updates, \model{} is significantly better than full softmax on the MIND memory
test in all three settings of \tabref{tab:ablation} (both data mixtures, both history modes), with significantly lower
MSE in each (not shown). Under an identical bounded KV budget, the hybrid raises PSNR by 0.89\,dB and is better in 44 of 50
segments, so the gain does not come from storing more history. With full history, it also completes the seven
segments (100--138\,s of prediction) on which full softmax runs out of GPU memory. On the recorded sets the same
ranking holds against ground truth at revisit instants (\tabref{tab:paired}, \appref{app:recorded}).

\begin{figure}[t]
\centering
\includegraphics[width=\linewidth]{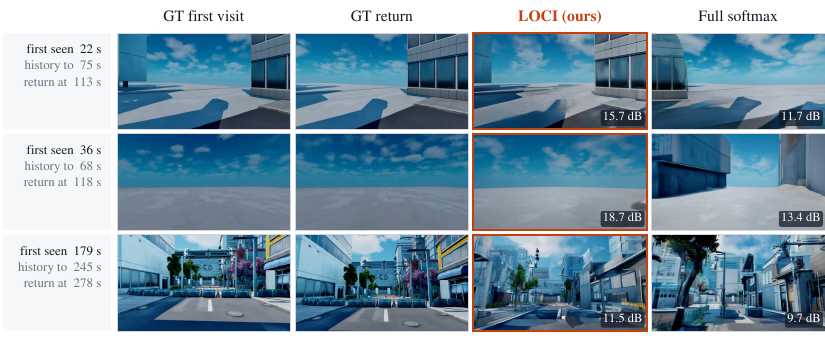}
\caption{\textbf{Long-range recall under the same bounded KV budget.} Both models receive the recorded video up to the stated time as history and generate the rest; shown are returns to places first seen more than 60\,s earlier, with the PSNR of the generated return frame against the corresponding ground truth. Full softmax, which can only attend to the retained frames, replaces the scene (new buildings, a tree), whereas \model{} restores it. We show three of the four trajectories, each at the return where \model{} leads by the largest margin.}
\label{fig:bounded_recall}
\end{figure}

\paragraph{Long-range recall under a bounded budget.}
To test recall well beyond the retained frames, we give both models the first 68--245\,s of four held-out recorded
trajectories (2 and 5\,min) as ground-truth history under the same bounded access, and let them generate the remaining
45--60\,s, which revisit places first seen at least 60\,s earlier. On all four trajectories, \model{} reproduces these
returns more faithfully than full softmax (revisit PSNR $+0.68$ to $+1.93$\,dB, lower LPIPS on each;
\figref{fig:bounded_recall}).

\paragraph{Direct training vs.\ conversion.}
Converting the trained full-softmax model into the same hybrid layout with the ARL$^2$ recipe does not reach the directly
trained hybrid, and neither converted model exceeds its own teacher (\appref{app:extra}).

\paragraph{Local consistency at short-horizon revisits.}
Frame-level averages dilute localized errors, such as an object that disappears or appears on return. At revisit instants
8--20\,s after the first visit, we therefore score the worst 5\% of patches by DINOv2 \citep{oquab2024dinov2} feature distance between the
generated and ground-truth revisit frames (\appref{app:localrevisit}). On the MIND memory test, this local error is
significantly lower for \model{} than for full softmax ($-0.073$) and lower in 12 of 13 segments with such revisits, whereas
whole-frame LPIPS on the same frames does not separate the models. The ranking holds at 1\% or 10\% and on a shared
worst region. Resetting the recurrent state of \model{} at every chunk during inference, all else unchanged, raises
this local error ($+0.021$ $[+0.013, +0.028]$) and blurs the revisit frames (Laplacian-variance
sharpness relative to ground truth 0.12 with the reset vs.\ 0.18 without), each in 15 of 16 segments. This comparison needs only \model{}'s own rollouts and therefore covers three more segments than the paired one. The recurrent state thus contributes to short-horizon fidelity
(examples in \figref{fig:local_revisit}, \appref{app:extra}). Per-token retention, trained under the same recipe (uniform mixture),
matches per-chunk retention at these short-range revisits; over all revisits of the MIND prediction segment, per-chunk
retention has lower local error, with the gap growing with the revisit interval (\appref{app:retention_ablation}).

\subsection{Analysis of the recurrent state}
\label{sec:analysis}
\label{sec:covis_attention}

\begin{figure}[t]
\centering
\includegraphics[width=0.8\linewidth]{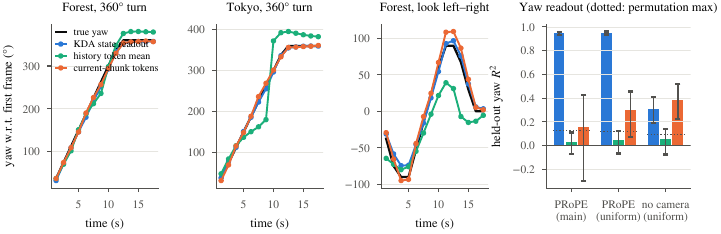}
\caption{Linear readout of camera yaw. Left three panels: true yaw and readouts from the KDA state, the mean of past
tokens, and the current chunk's input features (before camera conditioning) on example trajectories. Right: held-out $R^2$ (15-layer mean) for the main model (left group)
and, on the uniform-mixture branch with an identical recipe, the recurrent memory with PRoPE (middle) and without camera
encoding (right); dotted lines show the maximum over 20 label permutations.}
\label{fig:probe}
\end{figure}

\paragraph{Projective conditioning makes the recurrent state camera-decodable.}
PRoPE writes camera geometry into the recurrent keys and values. We test whether camera yaw relative to the first frame remains linearly decodable from the accumulated KDA state. We read the state at every chunk boundary from the ninth chunk on (32 designed trajectories; mean over the 15 hybrid layers; cross-validation and bootstrap intervals clustered by trajectory) and obtain held-out
$R^2=0.946$ $[0.929, 0.960]$ and a median error of 3.6$^\circ$ (\figref{fig:probe}); coordinate-invariant summaries of
the state (per-head singular values and norms) retain $R^2=0.941$. An untrained branch with the same conditioning decodes about as well, so decodability reflects the encoding rather than learned behaviour. Among models trained on the uniform mixture, PRoPE
raises this $R^2$ by $+0.646$ $[+0.544, +0.749]$ over a recurrent memory without camera encoding.

\paragraph{Historical attention concentrates on co-visible content, partly through the readout.}
We replay the same ground-truth history through both models and compare attention at the same query positions
(8 recorded clips, 154 revisit instants, token-level co-visibility from ground-truth depth used only for measurement;
\appref{app:covisdef}). Co-visible tokens make up only about 0.1\% of the history, yet the 15 softmax layers of \model{} place 11.1\% of their
history attention on them, versus 10.45\% for the same 15 layers of full softmax: an enrichment over uniform attention of $115\times$ vs.\
$108\times$ ($+6.64$), higher in all 8 clips. The difference persists when both models see the same retained
frames under bounded access ($88\times$ vs.\ $81\times$, $+6.82$), again in 8 of 8. Zeroing only the cross-chunk readout at inference
lowers the co-visible enrichment of the downstream softmax layers ($+0.56$ for on minus off), most strongly right after
a hybrid block ($+0.86$), and a norm-matched random readout lowers it further: downstream queries use the content of
the readout, although zeroing it removes only part of the difference to full softmax. On each model's own rollouts, \model{} also reconstructs the co-visible region more closely (masked LPIPS
0.597 vs.\ 0.654, lower at 131 of 154 instants), with the same sign in every tertile of co-visible attention
enrichment. Intervals for all these quantities are in \tabref{tab:covis}.

\subsection{Long-horizon generation and cost}
\label{sec:efficiency}

With full history, both models grow in memory and time per chunk until they exhaust the GPU.
Replacing half of the historical-KV layers with recurrent state extends the reachable length from
157\,s to 225\,s and lowers memory at equal length by about 30\% (\figref{fig:efficiency} and
\tabref{tab:cost} in \appref{app:extra}). \model{} is also faster than full softmax (7.2 vs.\ 8.1\,s per video second over
the first 40\,s), since half of its layers attend only within the current chunk. Under bounded sparse access, \model{} runs the full 300\,s at a constant
23.6\,GiB and constant time per chunk, and on the recorded set it scores slightly higher than with full
history (\tabref{tab:sparse}). This is 15\% less memory than full softmax under the same access (27.6\,GiB), at
the same speed: 5.4\,s per video second for both on one H200, with an output-preserving optimized implementation of
the recurrent branch (\appref{app:speed}).

%% file: sections/07_discussion.tex
\section{Discussion and Limitations}
\label{sec:disc}

\paragraph{Limitations.}
Our held-out trajectories are rendered in Unreal Engine; real-world captures are not evaluated. Absolute fidelity is low for all models (benchmark-average PSNR below 15\,dB), so the gains should be read as relative. We study one 5B backbone with a short fine-tuning budget (5,000 updates). The per-token retention comparison (Appendix~F) uses an extended metric defined after the short-range comparison. Both models keep an explicit-history camera-attention branch in all 30 blocks, so the recurrent path halves main-attention KV rather than all stored history. The recurrent state is built from noised chunk features in training but committed from generated chunks at inference. Speed parity with full softmax in the bounded sparse mode relies on the optimized implementation of Sec.~\ref{sec:efficiency}.
\paragraph{Conclusion.}
\model{} combines projective recurrent memory, chunk-level retention, and explicit
historical attention in a camera-controlled video model. The architecture keeps both
compact spatial context and direct access to retained visual detail, halves the number of main-attention layers that store historical KV, and supports dense and bounded
sparse historical access. On the MIND memory benchmark, it has the best MSE, PSNR and SSIM among the world models we evaluate
and improves over an identically trained full-softmax model under the same bounded KV budget. With a bounded bank of
retained observations, it streams at constant memory.

%% file: sections/08_appendix.tex
\section{Implementation Details}
\label{app:impl}

\subsection{Backbone and feature layout}
\label{app:layout}
The backbone has 30 transformer blocks, 24 heads per block, and head dimension 128.
The fifteen hybrid blocks have zero-based indices
$[2,4,6,7,8,9,11,13,14,16,23,24,25,27,28]$; the remaining blocks retain historical
softmax attention. At $512\times768$ image resolution, VAE latents have spatial size
$32\times48$ with 48 channels. Spatial patchification produces $16\times24=384$
tokens per latent frame. The temporal VAE stride is four. Training windows contain
$81=1+16\times5$ latent frames: one conditioning frame and sixteen future chunks.
Each hybrid block carries 24 recurrent states of size $128\times128$.

\begin{center}
\small
\begin{tabular}{lcc}
\toprule
Stored history & Full softmax & \model{} \\
\midrule
Main-attention KV layers & 30 & 15 \\
Fixed recurrent-state layers & 0 & 15 \\
Independent camera-attention layers & 30 & 30 \\
\bottomrule
\end{tabular}
\end{center}
The comparison assumes equal source-frame support, token resolution, and dtype; the camera
branch of every block stores the same retained frames as the historical-attention blocks. Weights, working activations,
current-chunk storage, and transfer buffers also occupy device memory, so this
layer count does not predict a factor-of-two reduction in total device usage.

\paragraph{Ray geometry.}
Camera poses are expressed relative to the first frame. For recurrent PRoPE, ray transforms
are constructed in FP32, their translation is divided by six, and they are composed with
normalized pinhole intrinsics. For image width $w$, height $h$, and horizontal field of
view $\theta_x$, the intrinsic construction is
\begin{equation}
 f_x=f_y=\frac{w}{2\tan(\theta_x/2)},\qquad
 (c_x,c_y)=(w/2,h/2),\qquad
 \overline K=\begin{bmatrix}
 f_x/w&0&c_x/w-1/2\\0&f_y/h&c_y/h-1/2\\0&0&1
 \end{bmatrix}.
\end{equation}
The homogeneous lift appends an identity coordinate. The translation divisor is an
encoding scale.
The independent UCPE camera-attention branch transforms its own projected queries,
keys, values and outputs using the ray-view matrices without this additional intrinsic
composition or translation divisor. Its rays still depend on the supplied camera intrinsics.

\paragraph{Head coordinates.}
The recurrent camera slot is $12{:}44$: four consecutive four-dimensional projective
tiles occupy $12{:}28$, followed by horizontal and vertical patch rotations on
$28{:}36$ and $36{:}44$. Each eight-dimensional patch rotation pairs the first four
coordinates with the last four. Channels $44{:}128$ use the backbone's 42 adjacent
spatial rotary pairs for queries and keys only. Channels $0{:}12$ use identity in the
recurrent branch. Writing the two patch rotations together as $R_i^p$ and the
backbone spatial rotation as $R_i^s$, the complete maps are
\begin{equation}
\begin{aligned}
 A_i&=I_{12}\oplus(P_i^\top)^{\oplus4}\oplus R_i^{p}\oplus R_i^{s},\\
 B_i&=I_{12}\oplus(P_i^{-1})^{\oplus4}\oplus R_i^{p}\oplus R_i^{s},\\
 C_i&=I_{12}\oplus(P_i^{-1})^{\oplus4}\oplus R_i^{p}\oplus I_{84}.
\end{aligned}
\label{eq:projective_maps}
\end{equation}
Here $\oplus$ denotes a block-diagonal sum. The inverse value map $C_i^{-1}$ decodes
the recurrent output. It differs from the inverse key map because values do not carry
the backbone spatial rotation.

\subsection{Gates and state execution}
The write gate is one scalar per token and head, $\beta=2\sigma(b(h))$.
For mean block input $\overline h_c$ and first token $i_c$, log-retention is
\begin{equation}
\begin{aligned}
 \gamma_c&=\max\!\left\{\log\alpha_{\min},
   -\exp(a)\,\operatorname{softplus}(W_\alpha\overline h_c+b_\alpha)\right\},\\
 D_i&=\begin{cases}\operatorname{diag}(\exp\gamma_c),&i=i_c,\\I,&i\ne i_c.\end{cases}
\end{aligned}
\label{eq:chunkdecay}
\end{equation}
The maximum is elementwise, $a$ is learned per head, and $\alpha_{\min}=0.2$.
In this equation, the projection and bias denote their expanded, channel-tied forms.
The retention projection produces 64 values per head, repeated over adjacent channel
pairs to form the 128 entries of $\gamma_c$ in \eqnref{eq:chunkdecay}. This parameter
sharing is retained as an implementation choice.
Initially the retention projection and $a$ are zero, with
$b_\alpha=\operatorname{softplus}^{-1}(-\log r_0)$ for an initial per-chunk retention $r_0=0.7$. Applying the log-retention floor once
at the start of a chunk gives a multiplicative retention of at least $0.2$ at that step;
all other steps use identity retention. Learned gates need not remain at their initial values.

The fused delta scan uses a key-by-value state and the update order in
\eqnref{eq:geokda}: decay first, residual correction second. Each chunk's readouts are
computed from its incoming state before this scan, rather than using the kernel's
within-chunk output sequence. Final states are stored in FP32. In the standard read path,
they are cast to the query dtype for the read operation; storing FP32 state does not make
the entire readout FP32.

During denoising, temporary state updates operate on a copy and are discarded.
Only the separate commit forward after sampling (timestep zero with full history; $t=100$ in the
bounded sparse setting) commits persistent state.
Classifier-free guidance maintains separate state streams for its two conditioning
branches. Explicit main and camera caches have a mutable current-chunk tail: its keys
and values are refreshed during denoising, then refreshed again from the final generated
latent. This tail is distinct from the protected recurrent state. Training uses a
forward-local scan of noised features and no additional clean-history commit pass.

\subsection{Execution schedule}
\label{app:state_schedule}
The schedule below applies independently to each hybrid layer. A single forward
passes token features through the interleaved network blocks; recurrent state is not
passed directly between layers. Historical softmax blocks read their retained KV using
queries projected from the current layer's input representation.

\begin{center}
\small
\begin{tabular}{p{0.20\linewidth}p{0.73\linewidth}}
\toprule
Stage & Recurrent-state operation \\
\midrule
Training input & Keep the first condition latent clean; independently noise each future
chunk. Start a forward-local state at zero. \\
Training forward & In causal chunk order, read the incoming state for all current queries,
then scan token-level writes to supply the next chunk's state. Discard this local state
after the forward; no separate clean commit is performed. \\
Inference prefill & Process the supplied condition at noise level zero to initialize
persistent state. \\
Denoising a chunk & Read the same committed incoming state at every noise level.
Temporary write results do not replace persistent state. \\
After sampling & If another chunk follows, run the final generated latent through a
new forward (timestep zero with full history; diffusion time $t=100$, noise level $0.1$, in the bounded sparse setting; the conditioning frame uses $t=0$) and commit its token-scan state. Refresh explicit current-chunk
KV from the same generated latent. \\
\bottomrule
\end{tabular}
\end{center}
Each forward propagates its output through ordinary transformer representations.
Consequently, a historical block downstream of a hybrid block can use recurrent context
in its queries without a dedicated query-conditioning module.

\subsection{Panorama-bank selection}
\label{app:bank}
Each candidate frame $j$ is represented by its camera center $p_j$ and a binary
field-of-view mask $M_j$ over 4,096 fixed Fibonacci-sphere directions. A direction is
included when it lies within the camera's rectangular pinhole field of view. This
orientation mask does not estimate occlusion or scene depth. For a frame $i$ and a set $S$,
let $\mathrm{cov}(i,S)=|M_i\cap\bigcup_{j\in S,\,\|p_i-p_j\|_2<6} M_j|/|M_i|$ be the fraction of
$i$'s directions already seen from retained frames within 6\,m. The five frames leaving the
recent window are visited in source order. A departing frame $f$ joins the bank only if
$1-\mathrm{cov}(f,\{0\}\cup\mathcal B)\ge0.3$, i.e.\ it adds at least 30\% new viewing
directions. When the bank then exceeds 20 frames, the bank frame with the largest
$\mathrm{cov}(b,\{0\}\cup(\mathcal B\setminus\{b\}))$ (the most redundant) is evicted, ties removing the newer
frame. The rule therefore favors earlier observations that add complementary coverage; the bank is
stored in source order. The initial condition is excluded because it has its own sink slot,
but it counts toward coverage. At each update at most the previous 20 bank frames plus five
departing recent frames are considered. Retention is coverage-based.

For a chunk starting at latent index $u_c$ with $h$ retained history frames
(sink, bank and recent window, in source order), the $i$-th retained frame
($i=0,\dots,h-1$) is assigned main-key time $u_c-h+i$: the retained frames are renumbered
compactly immediately before the current chunk, so the recent window keeps its source
times and the sink and bank occupy the preceding slots. Current frames retain source times. Camera attention uses the selected observations' original poses.
Main-key time correction applies the target rotary matrix multiplied by the inverse
source rotary matrix to the temporal channels $0{:}44$, using the actual stored
rotary coefficients. This preserves the distinction between native main-attention
coordinates and the recurrent coordinates of \eqnref{eq:projective_maps}. The remaining
main-key channels, cached values, and source camera poses are unchanged. The correction
is made on the read representation rather than repeatedly rotating canonical cached keys.
Retained-frame storage is bounded, whereas maximum executable duration also depends on
positional tables and the inference implementation.

\subsection{Scope of the streaming counts}
\label{app:streaming_counts}
For a fixed number of spatial tokens per frame, dense main-KV storage grows with the
retained prefix in all historical-attention layers: 30 for full softmax and 15 for
\model{}. Each of the 15 recurrent layers instead stores a fixed $24\times128\times128$
state per conditioning stream.

In sparse mode, the sink, bank, recent window, and current chunk together occupy at most
34 latent-frame slots in both explicit branches. The bank selects from at most 25
candidates per update, while recurrent state continues independently. For fixed chunk
size and denoising schedule, these history operations have bounded work per generated
chunk. A $C$-chunk rollout repeats them $C$ times; output storage and positional tables
are separate from the retained-history count.

\subsection{Method details: projective encoding, retention, and streaming storage}
\label{app:method_details}

\paragraph{Projective encoding.}
The maps $A_i$ and $B_i$ also retain the backbone's spatial rotary encoding
(\eqnref{eq:projective_maps}). Values are not L2-normalized; the readout is transformed
back by $C_i^{-1}$, including the inverse patch rotations. The full-head normalization in
\eqnref{eq:geokeys} is applied after the geometric transforms. Together with the state
update, it makes the recurrent memory a learned geometry-conditioned mechanism rather than an
invariant projective kernel. Temporal rotary encoding is disabled only in the KDA branch:
intra-chunk attention and the historical softmax blocks retain the backbone's native
temporal encoding, and the model remains causal. The independent camera-attention branch
retains its own ray-based transform.

\paragraph{Read and fusion.}
In \eqnref{eq:read}, head concatenation before the output projection $W_o$ is implicit,
and the gate adds the unnormalized memory readout to the local output. Each hybrid block
has its own state; the explicit bank stores observation features.

\paragraph{Retention once per chunk.}
Applying decay at every spatial token would make the number of forgetting steps depend
on tokenization as well as elapsed video time. Within a chunk of $N$ tokens, the content
written at position $i$ has been attenuated by the $N-i$ later gates when the chunk is
committed. Tokens of the same frame, which are observed simultaneously, are therefore
retained unequally (by a factor $r^{HW-1}$ between the first and last token of a frame, and
$r^{N-1}$ across the chunk). With the default per-token retention this leaves almost only
the final tokens of the last frame in the state. Even when the per-token rate is chosen
so that the product over a chunk matches ours, the first frame of a chunk retains only a
fraction $r_c^{(F-1)/F}$ of what the last frame retains ($F$ frames per chunk, $r_c$ the
per-chunk retention). Computing one retention per chunk from the chunk mean
(\eqnref{eq:chunkdecay}) removes this dependence for explicit retention: all tokens of an
observation share the same retention factor, and explicit forgetting advances only with
elapsed chunks. The delta-rule correction itself remains order-dependent within a chunk.

\paragraph{Commit pass.}
The commit forward recomputes the features of the final generated latent chunk; in the
bounded sparse setting it runs at diffusion time $t=100$ (noise level $0.1$) while the
conditioning frame still uses $t=0$. This pass uses the generated sample, not future
ground truth. The terminal chunk needs no commit when there is no subsequent query.

\paragraph{Bank and temporal indexing.}
The bank selects from the bounded candidate set of \eqnref{eq:recall}; a discarded
observation cannot later be retrieved from an external archive. Main historical keys use
a bounded temporal indexing scheme for the bank and sink, while camera attention preserves
each selected source pose. In the dense mode, camera history covers the same prefix as
main attention. We evaluate visual consistency separately from the
storage bound.

\paragraph{Streaming resource growth.}
\tabref{tab:streaming_storage} separates fixed recurrent state from explicit history.
The sparse
policy bounds explicit support in both architectures; recurrent memory adds a compressed
path for carrying historical context beyond that support. Let $T$ count latent frames
through the current chunk.
With token resolution, model width, chunk size, and denoising steps held fixed, dense
historical attention reads a growing prefix at each chunk even in the hybrid model. Sparse
access bounds both KV supports to at most 34 latent frames, including the current chunk,
and recurrent read/write work per chunk is independent of elapsed history.

\begin{table}[h]
\caption{\textbf{Logical history storage.} KV entries denote layers $\times$ retained latent-frame
support, not bytes; sparse entries are upper bounds. Each recurrent layer contains a
fixed set of per-head matrices. These are architecture counts, not measured device-memory ratios.}
\label{tab:streaming_storage}
\centering\scriptsize
\setlength{\tabcolsep}{5pt}
\renewcommand{\arraystretch}{1.1}
\begin{tabular}{lccc}
\toprule
\textbf{Mode} & \textbf{Recurrent state} & \textbf{Main KV} & \textbf{Camera KV} \\
\midrule
Full softmax, dense & None & $30\times T$ & $30\times T$ \\
\rowcolor{ourrow} \model{}, dense & 15 fixed layer states & $15\times T$ & $30\times T$ \\
Full softmax, sparse & None & $30\times\min(T,34)$ & $30\times\min(T,34)$ \\
\rowcolor{ourrow} \model{}, sparse & 15 fixed layer states & $15\times\min(T,34)$ & $30\times\min(T,34)$ \\
\bottomrule
\end{tabular}
\end{table}

Over $C$
chunks, sparse attention and recurrent work accumulate linearly under these fixed settings.
Dense prefix attention instead accumulates quadratic history-access work when prefix
length grows with $C$. End-to-end latency and peak device memory also depend on cache
placement, transfers, and other model operations, and are measured separately.

%% file: sections/09_appendix_protocol.tex
\section{Evaluation Protocol}
\label{app:bench}

\paragraph{Data, time, and model provenance.}
Training windows contain 81 latent frames and 321 RGB frames, sampled at
stride two from 24-fps recordings: about 26.67 seconds of source time, or
about 20 seconds at 16-fps playback. Unless stated otherwise, evaluation durations
(generated length, retained history and revisit gaps) are seconds at 16\,fps: one latent frame spans 0.25\,s and a
chunk of five latents 1.25\,s. MIND durations follow the benchmark's own 24-fps clock. Each training and recorded
evaluation clip is conditioned on a caption of its first frame produced by Qwen3-VL-8B \citep{qwen3vl2025}; MIND
clips use the benchmark's own descriptions.

Both \model{} and the full-softmax baseline start from the official pretrained Wan2.2-TI2V-5B backbone and are
trained for 5{,}000 updates with an identical recipe: global batch 64, 81-latent training windows, text dropout 0.1,
AdamW with a backbone learning rate of $10^{-5}$ and a camera-branch rate of $5\times10^{-5}$, a 250-update linear
warmup followed by cosine decay to $0.1\times$ over the 5{,}000 updates, zero weight decay, and gradient clipping at 0.3.
\model{} also uses a $10^{-4}$ learning rate for the recurrent branch. From update 2{,}000 onwards, both models
are trained on the same UE-weighted data mixture.

\paragraph{Training data and schedule.}
The main checkpoints branch at update 2{,}000 onto a UE-weighted data mixture and continue to 5{,}000
updates; the uniform-mixture branch is reported as an ablation in \secref{sec:ablation}.
Training uses 81-latent windows (about 20\,s at 16\,fps) drawn from Unreal Engine scenes that we captured and rendered (15{,}397 windows from 57 maps, 92.8\,h of source video), CARLA \citep{dosovitskiy2017carla} towns that we rendered (965 windows, 4 maps, 5.4\,h), and real walking videos (the Sekai-Walking subset \citep{li2025sekai} as processed by SolarWM \citep{solarwm2026}; no MIND clips). The first 2{,}000 updates and the uniform-mixture branch use 29{,}749 windows, 45\% of them real video; the UE-weighted branch uses 18{,}180 windows, 10\% of them real video (1{,}818 windows).
The recurrent branch adds 90.7M parameters (1.7\% of the 5.38B total); the full-softmax model has the same
backbone and camera branch without it.

\paragraph{Test-set details.}
No source episode of any test clip is used in training. The Forest Gas Station and Tokyo maps and the two maps of
the newer UE clips do not appear in training; Hwaseong and Temple Plaza are training maps traversed along new
trajectories.
Set A comprises all 18 held-out 40\,s walk-through clips from six scenes, with no appearance
filtering; each clip turns $570$--$950^\circ$ in total and repeatedly returns to earlier viewpoints, and
16 of the 18 clips contain the 204 revisit instants. Set B adds five held-out 40\,s clips selected for
frequent returns, with 59--86 revisit instants each. Set C contains 32 designed camera paths (static,
straight, spin, look-around, look-around with pitch, out-and-back, square) of 20--40\,s from four start
frames, without ground truth.

\paragraph{Camera following.}
\label{app:camera}
Camera following is a validity check rather than a contribution: it verifies that revisit fidelity is obtained while the
commanded trajectory is executed. \tabref{tab:camera} reports, on the 32 designed paths of set C, the median rotation error,
the median translation-direction error and the ratio of generated to commanded motion magnitude. We estimate them from the
generated videos by two-view geometry (SIFT matching and an essential matrix fitted with RANSAC in OpenCV, using the
known intrinsics).

\begin{table}[h]
\caption{\textbf{Camera following on the designed paths of set C} (medians). Mag.: generated / commanded motion magnitude. ---: the translation scale of AlayaWorld's outputs could not be calibrated.}
\label{tab:camera}
\centering\scriptsize
\setlength{\tabcolsep}{5pt}
\renewcommand{\arraystretch}{1.1}
\begin{tabular}{l ccc}
\toprule
Method & Rot.\,($^\circ$)$\downarrow$ & Trans.\,($^\circ$)$\downarrow$ & Mag.$\to$1 \\
\midrule
CaR \citep{car2026} & 2.14 & 8.0 & 0.74 \\
HY-WorldPlay \citep{hyworld15} & 2.75 & 9.6 & 1.38 \\
Matrix-Game 3.0 \citep{matrixgame3} & 1.33 & 6.8 & 2.09 \\
LingBot-World \citep{lingbotworld} & 0.72 & 4.9 & 1.06 \\
AlayaWorld \citep{alayaworld2026} & 7.49 & --- & --- \\
SANA-WM \citep{zhu2026sanawm} & 1.71 & 4.5 & 3.02 \\
SolarWM \citep{solarwm2026} & 1.13 & \textbf{3.8} & 3.08 \\
\addlinespace[2pt]
Full softmax (same recipe) & 0.38 & 4.7 & 1.05 \\
\rowcolor{ourrow} \textbf{\model{} (ours)} & \textbf{0.35} & 4.6 & \textbf{1.01} \\
\bottomrule
\end{tabular}
\end{table}

\paragraph{Dense and sparse access.}
Dense attention accesses the available generated prefix. The camera branch of every block uses the
same retained frames as the historical-attention blocks. Sparse access retains a protected conditioning
observation, recent context, and a bounded bank of older observations. A fixed-capacity bank
differs from a growing archive from which any old observation can later be recovered.
The evaluation protocol records selected source identities, bank capacity, recent-context length,
camera support, and the positional treatment of retained tokens. The primary two-model,
two-policy comparison uses identical 40-second trajectories; we report 300-second sparse rollouts in \secref{sec:efficiency}. Dense and sparse settings
preserve each model's checkpoint and camera encoding.
For the measured sparse setting, the maximum active support is 34 latent frames:
one conditioning frame, 20 bank frames, 8 recent frames, and five current frames.
The field-of-view coverage rule of \appref{app:impl} updates the bank: among at most 20 old bank
frames plus five departing recent frames, it favors earlier observations that add
complementary coverage. Generated frames are committed to history with a single forward at
diffusion time $t=100$ (noise level $0.1$); the conditioning frame is committed at $t=0$.
Both models use the same selected source identities in all 30 camera-attention layers.
At current-chunk start index $s$ with $h$ retained history frames, retained main keys are
renumbered compactly to positions $s-h,\dots,s-1$ in source order; recent keys therefore
retain their source positions.
This read-time temporal correction changes neither retained values nor source camera
poses. Main KV is host-offloaded for all 30 historical-softmax layers of the baseline
and the 15 historical-softmax layers of \model{}, with a 12-GiB GPU staging budget.
Camera KV is not host-offloaded.

\paragraph{Co-visible attention estimator.}
\label{app:covisdef}
\label{app:attention_scope}
Let $a_{\ell q k}$ denote attention weights averaged over heads before normalization, for
layer $\ell$, query $q$, and key $k$. With historical keys $\mathcal H_q$ and geometrically
co-visible historical keys $\mathcal V_q\subseteq\mathcal H_q$, the within-history mass is
\begin{equation}
 m_{\ell q}=\frac{\sum_{k\in\mathcal V_q}a_{\ell qk}}
 {\sum_{k\in\mathcal H_q}a_{\ell qk}}.
 \label{eq:covis_mass}
\end{equation}
Queries with zero historical mass are excluded and counted. Query means are aggregated
over the same historical-softmax layers in both models, followed by a case mean with its
number of probes disclosed. Averaging heads before normalization yields a
history-mass-weighted mixture of their conditional allocations. Co-visibility is computed at token level from ground-truth depth, which is used only for measurement and never for
generation; clips without depth use a finite-ray proxy and are reported separately.
For each probe we also record the total historical attention mass
$h_{\ell q}=\sum_{k\in\mathcal H_q}a_{\ell qk}$ and the co-visible candidate fraction
$\rho_q=|\mathcal V_q|/|\mathcal H_q|$. The former distinguishes a high conditional
co-visible allocation from strong use of history overall; the latter describes the
available geometric support. Ground-truth history is replayed teacher-forced and attention is read at one denoising
step ($t=500$) for all tokens of the revisit latent, in the 15 softmax layers of each model. Clips are aggregated with
equal weight per revisit instant and intervals are bootstrapped over clips.
\tabref{tab:covis} lists the results summarized in \secref{sec:analysis}.

\begin{table}[h]
\caption{\textbf{Co-visible attention and reconstruction} (8 recorded clips, 154 revisit instants; 95\% bootstrap
intervals over clips). Enrichment: co-visible share of history attention divided by co-visible share of historical tokens.}
\label{tab:covis}
\centering\scriptsize
\setlength{\tabcolsep}{5pt}
\renewcommand{\arraystretch}{1.1}
\resizebox{\linewidth}{!}{\begin{tabular}{l l c c}
\toprule
Comparison & Quantity & Value [95\% CI] & Count \\
\midrule
\model{} vs.\ full softmax, full history & co-visible attention share & 11.1\% vs.\ 10.45\% & \\
 & $\Delta$ enrichment & $+6.64$ [$+3.53$, $+8.45$] & 8 / 8 clips \\
\model{} vs.\ full softmax, bounded & co-visible attention share & 9.66\% vs.\ 8.91\% & \\
 & $\Delta$ enrichment & $+6.82$ [$+4.67$, $+7.72$] & 8 / 8 clips \\
Readout on $-$ off (\model{}) & $\Delta$ enrichment, downstream softmax layers & $+0.56$ [$+0.21$, $+0.77$] & \\
 & $\Delta$ enrichment, right after a hybrid block & $+0.86$ [$+0.45$, $+1.04$] & \\
\model{} vs.\ full softmax, own rollouts & masked LPIPS & 0.597 vs.\ 0.654 & \\
 & $\Delta$ masked LPIPS & $-0.057$ [$-0.113$, $-0.024$] & 131 / 154 instants \\
\bottomrule
\end{tabular}}
\end{table}

\section{Comparison on held-out recorded trajectories}
\label{app:recorded}

\begin{table}[h]
\caption{\textbf{Comparison with camera-controlled world models.} Revisit: generated vs.\ ground-truth
frames at the revisit instants of 20 clips (16 from set A, 4 from set B). Whole clip: all frames of set A (18 clips).
Identity: DINOv2 cosine between revisit and earlier views (set B). Camera following is checked in \appref{app:camera}.
Ours (full history and bounded sparse) and full softmax average three seeds on set A. Cost: s/v-s is seconds per video
second; for our models and full softmax, measured over the first 40\,s of a 300\,s trajectory on one H200 with the
output-preserving implementation of \appref{app:speed}. Best in \textbf{bold}.}
\label{tab:sota}
\centering\scriptsize
\setlength{\tabcolsep}{3.5pt}
\renewcommand{\arraystretch}{1.1}
\resizebox{\textwidth}{!}{%
\begin{tabular}{l cc cc ccc c cc}
\toprule
\multirow{2}{*}{\textbf{Method}} & \multirow{2}{*}{\textbf{Params}} & \multirow{2}{*}{\textbf{Steps}} & \multicolumn{2}{c}{\textbf{Revisit (A+B)}} & \multicolumn{3}{c}{\textbf{Whole clip (A)}} & \textbf{Identity} & \multicolumn{2}{c}{\textbf{Cost}} \\
\cmidrule(lr){4-5}\cmidrule(lr){6-8}\cmidrule(lr){9-9}\cmidrule(l){10-11}
& & & PSNR$\uparrow$ & LPIPS$\downarrow$ & PSNR$\uparrow$ & SSIM$\uparrow$ & LPIPS$\downarrow$ & DINO$\uparrow$ & s/v-s$\downarrow$ & GiB$\downarrow$ \\
\midrule
CaR \citep{car2026} & 5B & 50 & 9.73 & 0.627 & 8.89 & 0.360 & 0.647 & 0.679 & 31.4 & 17.7 \\
HY-WorldPlay \citep{hyworld15} & 8B & 4 & 10.01 & 0.640 & 9.70 & 0.408 & 0.625 & 0.519 & 14.7 & 71.1 \\
Matrix-Game 3.0 \citep{matrixgame3} & 5B & 3 & 8.40 & 0.687 & 7.85 & 0.313 & 0.678 & 0.626 & 6.2 & 35.0 \\
LingBot-World \citep{lingbotworld} & 28B & 4 & 8.70 & 0.605 & 9.67 & \textbf{0.442} & 0.564 & 0.517 & 20.5 & 88.3 \\
AlayaWorld \citep{alayaworld2026} & 15B & 4 & 8.52 & 0.646 & 8.34 & 0.386 & 0.640 & -- & 24.9 & 137.6 \\
SANA-WM \citep{zhu2026sanawm} & 2.6B & 4 & 9.63 & 0.771 & 9.41 & 0.265 & 0.752 & 0.449 & \textbf{2.4} & 49.0 \\
SolarWM \citep{solarwm2026} & 5B & 4 & 7.78 & 0.683 & 8.04 & 0.341 & 0.646 & 0.459 & 8.2 & 36.1 \\
\addlinespace[2pt]
Full softmax (same recipe) & 5B & 50 & 10.02 & 0.579 & 10.28 & 0.415 & 0.578 & 0.659 & 8.1 & 44.2 \\
\rowcolor{ourrow} \textbf{\model{} (ours)} & 5B & 50 & 10.64 & 0.561 & 10.63 & 0.414 & 0.571 & \textbf{0.796} & 7.2 & 33.6 \\
\rowcolor{ourrow} \model{} (bounded sparse) & 5B & 50 & \textbf{11.21} & \textbf{0.547} & \textbf{10.86} & 0.441 & \textbf{0.559} & 0.639 & 5.4 & \textbf{23.4} \\
\bottomrule
\end{tabular}}
\end{table}

\begin{table}[h]
\caption{\textbf{Paired revisit comparison.} \model{} minus the other method on the same 20 clips,
with 95\% bootstrap intervals; $\Delta$LPIPS is other $-$ ours, so positive favors \model{}.
\emph{Ours lower} counts clips with lower LPIPS for \model{}. Bold: interval excludes zero.}
\label{tab:paired}
\centering\scriptsize
\setlength{\tabcolsep}{5pt}
\renewcommand{\arraystretch}{1.1}
\begin{tabular}{l r@{\;}l r@{\;}l c}
\toprule
\multirow{2}{*}{\textbf{Versus}} & \multicolumn{2}{c}{\textbf{$\Delta$PSNR}$\uparrow$} & \multicolumn{2}{c}{\textbf{$\Delta$LPIPS}$\uparrow$} & \multirow{2}{*}{\textbf{Ours lower}} \\
\cmidrule(lr){2-3}\cmidrule(lr){4-5}
 & Mean & 95\% CI & Mean & 95\% CI & \\
\midrule
Full softmax (same recipe) & \textbf{+0.62} & {\tiny [+0.13, +1.12]} & +0.018 & {\tiny [$-$0.006, +0.042]} & 14 / 20 \\
HY-WorldPlay \citep{hyworld15} & +0.63 & {\tiny [$-$0.49, +1.71]} & \textbf{+0.079} & {\tiny [+0.037, +0.122]} & 14 / 20 \\
CaR \citep{car2026} & +0.91 & {\tiny [$-$0.10, +2.16]} & \textbf{+0.066} & {\tiny [+0.007, +0.128]} & 13 / 20 \\
SANA-WM \citep{zhu2026sanawm} & \textbf{+1.02} & {\tiny [+0.03, +2.04]} & \textbf{+0.210} & {\tiny [+0.146, +0.276]} & 19 / 20 \\
LingBot-World \citep{lingbotworld} & \textbf{+1.95} & {\tiny [+0.96, +2.99]} & +0.044 & {\tiny [$-$0.027, +0.112]} & 12 / 20 \\
AlayaWorld \citep{alayaworld2026} & \textbf{+2.12} & {\tiny [+1.20, +3.19]} & \textbf{+0.085} & {\tiny [+0.021, +0.150]} & 15 / 20 \\
Matrix-Game 3.0 \citep{matrixgame3} & \textbf{+2.24} & {\tiny [+1.27, +3.26]} & \textbf{+0.126} & {\tiny [+0.080, +0.177]} & 18 / 20 \\
SolarWM \citep{solarwm2026} & \textbf{+2.86} & {\tiny [+0.62, +4.76]} & \textbf{+0.122} & {\tiny [+0.028, +0.219]} & 15 / 20 \\
\bottomrule
\end{tabular}
\end{table}

\tabref{tab:sota} and \tabref{tab:paired} summarize the comparison on the held-out recorded trajectories. At revisit instants, \model{} produces the
frames closest to the ground truth: its revisit LPIPS is significantly lower than that of six of the seven
external models, and its revisit PSNR significantly higher than that of five of them, although several of them are
larger (8--28B) and use retrieval-based memory. Across all frames of the unfiltered recorded set, \model{} has the
highest PSNR, with results stable across seeds ($\pm$0.14\,dB). Revisit frames are also recognized as the same place
most reliably (DINOv2 0.796 vs.\ at most 0.679 for external models; CLIP \citep{radford2021clip} 0.928 vs.\ at most 0.900).
As a validity check, \model{} executes the commanded camera paths (median rotation error 0.35$^\circ$, magnitude
ratio 1.01; \appref{app:camera}), so its revisit fidelity is not obtained by moving less than commanded. Against the full-softmax model trained with the same data and recipe, the hybrid improves revisit PSNR by
0.62\,dB (significant) and identity consistency from 0.659 to 0.796.
Bounded access scores higher than full history at revisits on these trajectories, whose history is generated from a
single frame, whereas full history scores higher on MIND, whose history is a long recorded segment. Training windows
cover at most 81 latent frames (about 20\,s); attending to the full self-generated history of a 40\,s rollout is
therefore further from the training distribution than the bounded set, while a recorded memory segment supplies clean
evidence. This likely explains the difference.

\section{MIND over the first 20 seconds}
\label{app:mind20}
\tabref{tab:mind20} repeats the comparison of \tabref{tab:thirdparty} over the first 20\,s of the prediction segment
(480 frames at 24\,fps) for the models we run, with the same outputs and evaluation code.

\begin{table}[h]
\caption{\textbf{MIND memory test, first 20\,s of the prediction segment.} Same models, outputs and scorer as
\tabref{tab:thirdparty}.}
\label{tab:mind20}
\centering\scriptsize
\setlength{\tabcolsep}{4.5pt}
\renewcommand{\arraystretch}{1.1}
\begin{tabular}{l c cccc}
\toprule
Method & $n$ & MSE$\downarrow$ & PSNR$\uparrow$ & SSIM$\uparrow$ & LPIPS$\downarrow$ \\
\midrule
HY-WorldPlay \citep{hyworld15}       & 50 & 0.0672 & 12.84 & 0.411 & 0.695 \\
Matrix-Game 3.0 \citep{matrixgame3}  & 49 & 0.0737 & 12.51 & 0.387 & \textbf{0.635} \\
AlayaWorld \citep{alayaworld2026}   & 50 & 0.0815 & 11.97 & 0.379 & 0.659 \\
Alaya-EVOKE \citep{evoke2026}       & 50 & 0.0814 & 11.65 & 0.369 & 0.687 \\
LingBot-World \citep{lingbotworld}   & 50 & 0.0855 & 11.96 & 0.363 & 0.669 \\
\rowcolor{ourrow} \textbf{\model{} (ours)}  & 50 & \textbf{0.0478} & \textbf{14.12} & \textbf{0.454} & \textbf{0.635} \\
\rowcolor{ourrow} \model{} (bounded sparse) & 50 & 0.0601 & 13.34 & 0.432 & 0.661 \\
\bottomrule
\end{tabular}
\end{table}

\subsection{How each external baseline receives the MIND memory segment}
\label{app:mindadapters}
Every external method in \tabref{tab:thirdparty} and \tabref{tab:mind20} is given the same memory segment (the raw
MIND frames before \texttt{mark\_time}) and the same target camera path (the continuous ground-truth pose of the
prediction segment) as \model{}. Each adapter subclasses or wraps the official inference pipeline without modifying
it, and feeds the memory segment through whichever mechanism the official code already exposes for consuming its own
generation history (\tabref{tab:mindadapters}). The reference rows of \tabref{tab:thirdparty} (SSM, FramePack,
Context-as-Memory, GIM-World) are copied from Table~1 of the GIM-World paper \citep{gimworld2026}, where all four were
trained on MIND data with a Wan2.1 backbone; they are not run through an adapter and are therefore not listed below.

\begin{table}[h]
\caption{\textbf{How each external baseline receives the MIND memory segment.} Res.\ / fps: resolution and frame
rate of the official pipeline's own generation. Steps / CFG: sampling steps and classifier-free guidance scale used
(distilled checkpoints run CFG-free). No official inference code was modified: every baseline runs through an adapter
that subclasses or wraps it unmodified.}
\label{tab:mindadapters}
\centering\scriptsize
\setlength{\tabcolsep}{3pt}
\renewcommand{\arraystretch}{1.2}
\begin{tabular}{L{1.55cm} L{5.5cm} L{2.75cm} L{1.85cm} L{1.15cm}}
\toprule
\textbf{Method} & \textbf{Memory-segment feed} & \textbf{Control signal} & \textbf{Res.\ / fps} & \textbf{Steps / CFG}\\
\midrule
HY-WorldPlay \citep{hyworld15} & Retrieval-augmented KV-cache: memory GT frames VAE-encoded to clean latents and
written into the latent tensor in place of the noise-initialized first chunk; later chunks retrieve by camera-FOV
overlap over the full latent history (avg.\ 1141 frames, range 255--2406) & Continuous GT pose per predicted frame
(no WASD discretization) & 832$\times$480, 24\,fps & 4 / none \\
Matrix-Game~3.0 \citep{matrixgame3} & Long-range FOV-retrieval bank (\texttt{x\_memory}, selected from the
official AR history tensors) plus a 16-frame local re-diffusion window; memory GT frames VAE-encoded to clean latents
and prefilled into the same history tensors (avg.\ 1144 frames, range 255--2406) & Discrete WSAD~+~mouse actions,
quantized from MIND's continuous GT poses (12.35\,cm/frame translation; yaw $\pm1.5^\circ$/frame) & 1280$\times$704
native (17\,fps) $\to$ 1280$\times$720 at 24\,fps for scoring & 3 / none \\
LingBot-World \citep{lingbotworld} & Rolling self-attention KV-cache (sink 6 + recent 12 latent frames); the full
memory segment is causally VAE-encoded chunk by chunk and written with the same clean-latent ($t{=}0$) forward pass
the model uses for its own history, but only the sink ($\approx$1.3\,s) and tail ($\approx$3\,s) remain in the
window & Continuous GT pose via Pl\"ucker camera embedding & 832$\times$464, 16\,fps & 4 / none \\
AlayaWorld \citep{alayaworld2026} & Sink (1 frame) + 4-latent history window + 9-frame RGB motion neighbors from the last 25 memory frames,
plus the entire memory segment written into a ViGeo spatial point-map bank (top-10 frames retrieved per step by
target-pose coverage) & Continuous GT camera pose (OpenCV c2w) & 544$\times$960, 24\,fps & 4 / 3.0 \\
Alaya-EVOKE \citep{evoke2026} & World-state-library prefill (v2v mode): the entire memory segment and its GT poses are fed to a
persistent point-cloud world-state library (per-chunk monocular-depth back-projection); read = pose-indexed
retrieval, top-8 covisible sources, z-buffer warp & Continuous GT camera pose & 384$\times$640, 24\,fps & 3 / none
(guidance 1.0) \\
CaR \citep{car2026} & Context frames: the entire memory segment (arbitrary length) is VAE-encoded and temporally
resampled onto the memory-token grid by the official memory encoder & Continuous GT camera pose & 480$\times$832,
24\,fps & 50 / 3.0 \\
\bottomrule
\end{tabular}
\end{table}

\section{Relation to hybrid recurrent video models}
\label{app:hybrid_related}

\begin{table}[h]
\caption{\textbf{Hybrid recurrent video models.} How each model is conditioned, which history its softmax attention
reaches, what its recurrent path is, and how memory is trained and evaluated.}
\label{tab:hybrid_related}
\centering\scriptsize
\setlength{\tabcolsep}{2.5pt}
\renewcommand{\arraystretch}{1.2}
\begin{tabular}{L{1.5cm} L{1.95cm} L{2.4cm} L{3.3cm} L{1.8cm} L{1.85cm}}
\toprule
 & \textbf{Task / conditioning} & \textbf{Softmax access} & \textbf{Recurrent path} & \textbf{Training} & \textbf{Memory evaluation} \\
\midrule
Video SSM \citep{po2025ssm} & actions (Memory Maze, Minecraft) & dense local frame attention & block-wise state-space scan (per the authors, trading spatial consistency for memory) & trained on game data & spatial retrieval on synthetic scenes \\
Hybrid Forcing \citep{li2026hybridforcing} & text-to-video & local window & additive linear state of evicted KV & dense-to-hybrid distillation & -- \\
ARL$^2$ \citep{li2026arl2} & text-to-video, no camera & current frame (hybrid layers); causal softmax (others) & gated delta rule with per-token gates & conversion by layer-output and velocity distillation & VBench \\
SANA-WM \citep{zhu2026sanawm} & single image + camera path & attention sink + local window & Gated DeltaNet, one recurrent step per latent frame; UCPE camera encoding & multi-stage training & revisit vs.\ another generated frame \\
\model{} (ours) & memory segment or first frame + camera path & all retained observations (full, or first + bank of 20 + 8 most recent) & KDA, token-wise writes, retention once per chunk, reads from the preceding state; PRoPE on $q,k,v$ & direct diffusion-forcing training from Wan2.2 & revisit vs.\ recorded ground truth \\
\bottomrule
\end{tabular}
\end{table}

\paragraph{Comparison with SANA-WM.}
SANA-WM interleaves frame-wise Gated DeltaNet with softmax blocks restricted to an attention sink and a local window,
so content outside the window is available only through the recurrent state. \model{} keeps observation-level history
directly accessible to softmax attention and uses the camera-conditioned recurrent readout as context for its queries.
On the recorded trajectories (\tabref{tab:sota}), where every model receives the same first frame and camera path,
\model{} is better at revisit instants in PSNR ($+1.02$\,dB $[+0.03, +2.04]$) and LPIPS (0.210 lower, $[0.146, 0.276]$),
with lower LPIPS in 19 of 20 clips, and has higher identity consistency (DINO 0.796 vs.\ 0.449). SANA-WM's released
pipeline accepts a single conditioning image, so on MIND it receives the last frame of the memory segment; over the
entire prediction segment it obtains 0.0856 / 11.18 / 0.345 / 0.743 (MSE / PSNR / SSIM / LPIPS), against
0.0455 / 14.36 / 0.464 / 0.643 for \model{}. SANA-WM is faster (2.4 vs.\ 7.2\,s per video second with full history and 5.4\,s with bounded access), with 2.6B
parameters and four distilled steps against our 5B and 50 steps.

\paragraph{Video SSM.}
Retrained on MIND data by the GIM-World authors, Video SSM obtains 0.0796 / 11.96 / 0.395 / 0.744
(\tabref{tab:thirdparty}); \model{}, not trained on MIND, obtains 0.0455 / 14.36 / 0.464 / 0.643.

\section{Additional results}
\label{app:extra}

\begin{table}[h]
\caption{\textbf{WBench-Navi gated camera-return consistency} (camera-controlled navigation cases whose trajectory
returns to an earlier view; higher is better). External models: the models evaluated in the WBench paper, with values from its official leaderboard. \model{}: official evaluation code on
the same cases.}
\label{tab:thirdparty2}
\centering\scriptsize
\setlength{\tabcolsep}{4pt}
\renewcommand{\arraystretch}{1.1}
\begin{tabular}{rlc rlc}
\toprule
\# & Model & Score & \# & Model & Score \\
\midrule
1 & HY-World 1.5 \citep{hyworld15} & 84.92 & 12 & Wan 2.7 & 71.00 \\
\cellcolor{ourrow}2 & \cellcolor{ourrow}\textbf{\model{} (ours)} & \cellcolor{ourrow}82.00 & 13 & LTX 2.3 & 70.17 \\
3 & Matrix-Game 3.0 \citep{matrixgame3} & 80.37 & 14 & LingBot-World \citep{lingbotworld} & 67.14 \\
4 & Genie 3 & 78.39 & 15 & InSpatio-World & 66.47 \\
5 & Happy Oyster & 75.83 & 16 & LongCat-Video & 66.23 \\
6 & Kling 3.0 & 75.14 & 17 & Matrix-Game 2.0 & 64.51 \\
7 & HY-Video 1.5 & 75.12 & 18 & Fantasy-World & 64.23 \\
8 & Infinite-World & 74.37 & 19 & Astra & 63.30 \\
9 & Cosmos 2.5 & 74.32 & 20 & Kairos 3.0 & 62.01 \\
10 & Seedance 1.5 & 72.42 & 21 & HY-GameCraft & 60.51 \\
11 & YUME 1.5 & 71.40 & & & \\
\bottomrule
\end{tabular}
\end{table}
\begin{table}[h]
\caption{\textbf{Peak GPU memory versus generated length} (300\,s trajectory, one H200; peak allocated, GiB).}
\label{tab:cost}
\centering\scriptsize
\setlength{\tabcolsep}{5pt}
\renewcommand{\arraystretch}{1.1}
\begin{tabular}{ll c ccc c}
\toprule
Model & History & Reached & 60\,s & 120\,s & 180\,s & Max \\
\midrule
Full softmax & full   & OOM at 157\,s & 63.6 & 102.8 & ---  & 128.8 \\
\rowcolor{ourrow} \model{}     & full   & OOM at 225\,s & 45.5 & 69.0  & 100.2 & 114.9 \\
Full softmax & sparse & 300\,s & 27.4 & 27.5 & 27.5 & 27.6 \\
\rowcolor{ourrow} \model{}     & sparse & 300\,s & 23.4 & 23.4 & 23.5 & 23.6 \\
\bottomrule
\end{tabular}
\end{table}
\begin{table}[h]
\caption{\textbf{Bounded sparse access vs.\ full history} on the recorded set A (16 clips with revisit instants;
generated vs.\ ground truth; seed 42). Sparse: first frame, a bank of 20 retained latents and the 8 most recent
latents (identical for both models), with the recurrent state of \model{} carried over the full history.}
\label{tab:sparse}
\centering\scriptsize
\setlength{\tabcolsep}{5pt}
\renewcommand{\arraystretch}{1.1}
\begin{tabular}{l cc cc c}
\toprule
& \multicolumn{2}{c}{\textbf{Revisit}} & \multicolumn{2}{c}{\textbf{Whole clip}} & \textbf{Peak memory} \\
\cmidrule(lr){2-3}\cmidrule(lr){4-5}\cmidrule(l){6-6}
Model, history access & PSNR$\uparrow$ & LPIPS$\downarrow$ & PSNR$\uparrow$ & LPIPS$\downarrow$ & (GiB, grows / constant) \\
\midrule
Full softmax, full history   & 10.16 & 0.576 & 10.18 & 0.587 & grows (OOM at 157\,s) \\
\rowcolor{ourrow} \model{}, full history       & 10.67 & 0.544 & 10.36 & 0.577 & grows (OOM at 225\,s) \\
Full softmax, bounded sparse & 10.15 & 0.570 & 10.07 & 0.578 & 27.6, constant \\
\rowcolor{ourrow} \model{}, bounded sparse     & 11.14 & 0.524 & 10.49 & 0.563 & 23.6, constant \\
\bottomrule
\end{tabular}
\end{table}

\begin{figure}[h]
\centering
\includegraphics[width=\linewidth]{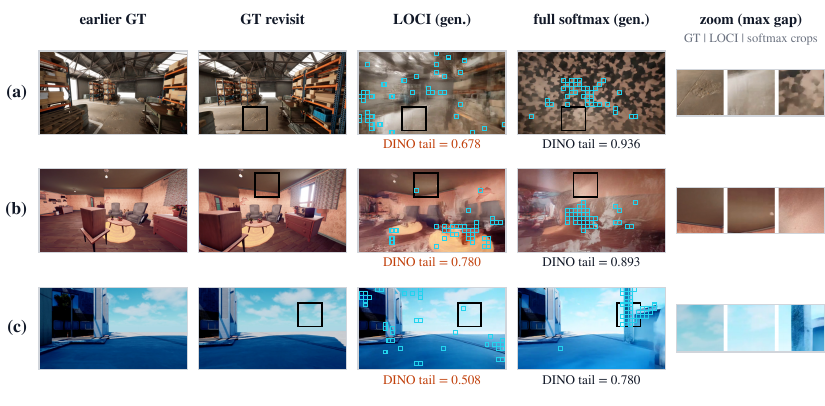}
\caption{\textbf{Localized consistency at short-horizon revisits.} Earlier ground-truth view, ground-truth revisit, and
the revisit generated by \model{} and by full softmax (same recipe); outlines mark each model's worst 5\% of patches by
DINOv2 distance to the ground truth, and the right column zooms into the region where the two models differ most.
Instants are chosen by a fixed rule (bright scenes, ranks 1, 3, 5 of the paired difference).}
\label{fig:local_revisit}
\end{figure}

\paragraph{Local-consistency metric.}
\label{app:localrevisit}
Frame-level averages dilute localized errors such as an object that disappears or appears on return, which occupy a
few percent of the frame. We therefore score, at revisit instants 8--20\,s after the first visit (defined from ground-truth
poses only), the worst 5\% of $14{\times}14$ patches by DINOv2 feature distance between the generated and the ground-truth
revisit frame, allowing a one-patch displacement (DINOv2-base, last layer, $37{\times}21$ patch grid; error $=1-$ the maximum cosine
similarity within $\pm1$ patch at the same location; each model's own worst patches). On the MIND memory test this local error is significantly lower for
\model{} than for full softmax (paired difference $-0.073$ $[-0.095, -0.050]$, 13 segments with such revisits, lower in 12), whereas
the whole-frame LPIPS of the same frames does not separate the models ($-0.007$ $[-0.033, +0.022]$); on the recorded sets
the local error is lower in 12 of 18 clips ($-0.020$ $[-0.047, -0.000]$). This ranking is unchanged when the worst fraction
is set to 1\% or 10\%, or when the two models are scored on a shared worst region. On the MIND memory test, the
local-vs-whole-frame gap is itself significant (difference of differences $+0.066$ $[+0.038, +0.095]$, clip-clustered
bootstrap): the hybrid's advantage over full softmax is concentrated in localized short-horizon fidelity.

\paragraph{Retention once per chunk vs.\ per token.}
\label{app:retention_ablation}
We train a variant with per-token retention (the KDA default) from the same initialization, with the same data
(uniform mixture), recipe and 5{,}000 updates as the chunk-level model, and compare the two with the same bounded
sparse access. Revisits 8--20\,s apart within the first 20\,s, as used above, probe only short-range memory and do not
separate the two ($-0.005$ $[-0.020, +0.011]$ on the recorded sets). We therefore extend the metric to every revisit
instant over the entire MIND prediction segment, including returns to the memory segment, with at most 60 instants per
segment stratified by gap (all 50 segments, 2{,}928 instants). We defined this extension after the short-range
comparison; it is exploratory. Chunk-level retention has lower local error, and the difference grows with the revisit
gap (\tabref{tab:retention_ablation}), consistent with forgetting that follows elapsed video time rather than token
position.

\begin{table}[h]
\caption{\textbf{Per-token minus per-chunk retention}: local DINOv2 error at MIND revisit instants over the entire
prediction segment, by revisit gap (positive: chunk-level lower; paired over segments, 95\% bootstrap intervals).}
\label{tab:retention_ablation}
\centering\scriptsize
\setlength{\tabcolsep}{5pt}
\begin{tabular}{l c c c}
\toprule
Revisit gap & Segments & $\Delta$ local error & Chunk lower \\
\midrule
8--20\,s  & 26 & $+0.0045$ [$-0.0013$, $+0.0103$] & 17 / 26 \\
20--60\,s & 44 & $\mathbf{+0.0101}$ [$+0.0039$, $+0.0169$] & 28 / 44 \\
$>$60\,s  & 35 & $\mathbf{+0.0126}$ [$+0.0038$, $+0.0210$] & 27 / 35 \\
All       & 50 & $\mathbf{+0.0084}$ [$+0.0035$, $+0.0133$] & 35 / 50 \\
\bottomrule
\end{tabular}
\end{table}

\paragraph{Retention after training.}
With one decay per chunk computed from the chunk's content, the per-chunk retention of \model{} varies with the input.
Training also separates
a group of long-memory channels in intermediate layers: after 5{,}000 updates, 6.4--11.1\% of the channels of layers
11, 13 and 23 have mean retention above 0.9 (half-life beyond six chunks), compared with at most 2.1\% after 1{,}000 updates.

\paragraph{Direct training vs.\ conversion.}
Converting the trained full-softmax model into the same hybrid layout with the ARL$^2$ recipe, on real data or on teacher
rollouts, does not reach the directly trained hybrid: on the recorded set A, \model{} has higher revisit PSNR
($+0.91$ $[+0.07, +1.74]$\,dB against conversion on real data) and lower revisit LPIPS than both converted models
($-0.030$ $[-0.055, -0.004]$ and $-0.050$ $[-0.088, -0.009]$), and neither converted model exceeds its own teacher. The
recurrent memory therefore needs to be trained jointly with camera-conditioned generation.

\begin{figure}[h]
\centering
\includegraphics[width=0.5\linewidth]{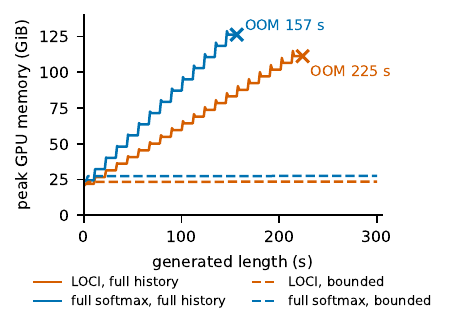}
\caption{\textbf{Cost versus generated length} on a 300\,s recorded trajectory (single H200, 141\,GiB).
Peak GPU memory. With full history both models grow until they
exhaust the GPU; bounded sparse access keeps memory and per-chunk time flat to 300\,s.}
\label{fig:efficiency}
\end{figure}

\paragraph{Inference implementation of the recurrent branch.}
\label{app:speed}
Per layer and denoising step, the recurrent read needs about 80$\times$ fewer FLOPs than softmax attention over the
retained history, but a direct implementation is dominated by the launch overhead of many small operations. Our
implementation builds the projective transform once per chunk and shares it across layers, skips the state update that
denoising steps compute and discard (only the separate commit pass writes the state), and captures the recurrent branch
of each layer in a CUDA graph. The final latents and all committed states are bit-identical to the direct
implementation. In the bounded sparse mode on one H200 this gives 5.43\,s per video second for \model{} and 5.45\,s for
full softmax. With full history over the first 40\,s, \model{} takes 7.2\,s per video second and 33.6\,GiB, against
8.1\,s and 44.2\,GiB for full softmax, since half of its layers attend only within the current chunk.

\paragraph{Cost of external models.}
Among the compared world models (\tabref{tab:sota} in \appref{app:recorded}), CaR requires 31.4\,s per video second and exceeds GPU memory on
long trajectories, and the larger models require 71--138\,GiB; HyDRA \citep{hydra2026} and SPMem \citep{longtermspatial2025}, which use
bidirectional attention within each generated clip, require 175 and 97\,s per video second.